\pdfoutput=1

\documentclass[11pt]{article}

\usepackage[preprint]{acl}

\usepackage{times}
\usepackage{paralist, tabularx}
\usepackage{latexsym}
\usepackage{graphicx}
\usepackage{multicol}
\usepackage{multirow}
\usepackage{amsmath}
\usepackage{booktabs}
\usepackage{subcaption}
\usepackage{soul}
\usepackage{color}
\usepackage{xcolor}

\usepackage{authblk}
\usepackage[T1]{fontenc}

\usepackage[utf8]{inputenc}

\usepackage{microtype}

\usepackage{inconsolata}

\usepackage{xspace}
\newcommand{\method}{\textsc{EDT}\xspace}
\usepackage{pifont}
\title{EDT: Improving Large Language Models' Generation by Entropy-based Dynamic Temperature Sampling}

\newcommand*{\affaddr}[1]{#1} 
\newcommand*{\affmark}[1][*]{\textsuperscript{#1}}

\author{
\textbf{Shimao Zhang}\affmark[$\dag$], \textbf{Yu Bao}, \textbf{Shujian Huang}\affmark[$\dag$]\thanks{Corresponding authors}\\
\affaddr{\affmark[$\dag$]National Key Laboratory for Novel Software Technology, Nanjing University}\\
\texttt{smzhang@smail.nju.edu.cn, nlp.baoy@gmail.com, huangsj@nju.edu.cn}
}

\begin{document}
\maketitle
\begin{abstract}
Recently, Large Language Models (LLMs) have demonstrated outstanding performance across a wide range of downstream language tasks. Temperature sampling is a commonly used decoding strategy for LLMs' generation process. However, a fixed temperature parameter is used in most cases, which may not always be an optimal choice for balancing generation quality and diversity. In this paper, we propose an effective Entropy-based Dynamic Temperature (EDT) Sampling method, to achieve a more balanced performance in terms of both generation quality and diversity by dynamically selecting the temperature parameter. Additionally, we also show model performance and comprehensive analyses for 4 different generation benchmarks. Our experiments show that EDT significantly outperforms the existing strategies across different tasks.
\end{abstract}

\section{Introduction}\label{sec:introduction}
Natural Language Generation (NLG) is an important part of Natural Language Processing (NLP), which aims to generate natural language content based on some provided textual inputs in a specific language task situation. Meanwhile, Large Language Models (LLMs) have been widely applied to natural language generation tasks~\citep{brown2020language, chowdhery2023palm, touvron2023llama}, achieving remarkable results in tasks such as question answering~\citep{zou2023representation}, summarization~\citep{pang2022long}, machine translation~\citep{zhu2023multilingual}, and more. The performance on these tasks demonstrates the impressive language capabilities of LLMs.

\begin{figure}[ht]
    \centering
    \includegraphics[width=0.48\textwidth]{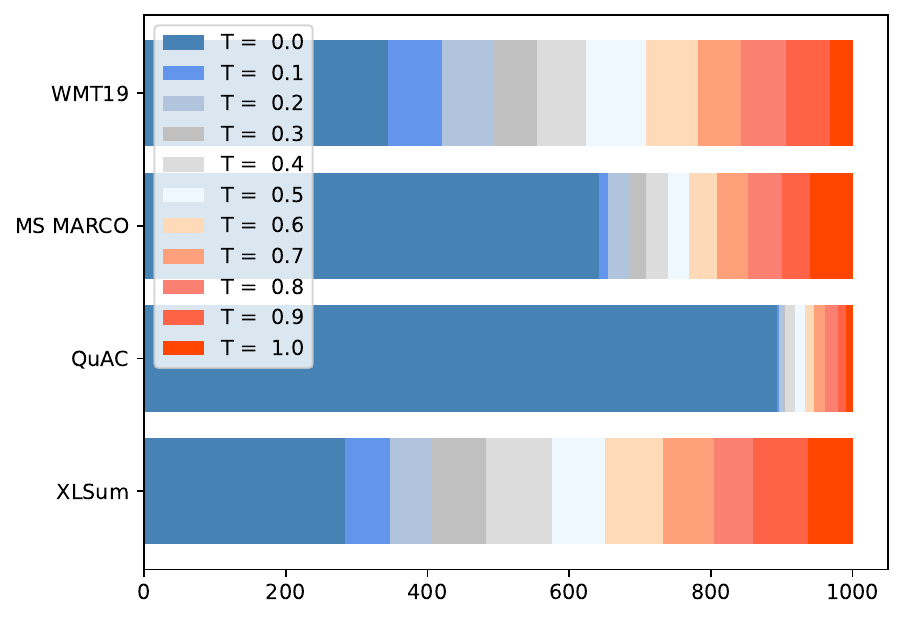}
    \caption{\textbf{Temperature distribution for optimal generation quality score on four datasets at single instance level.} The horizontal axis represents the number of instances. All experiment settings follow the same settings in Section \ref{sec:experiments}. This result is discussed in \S \ref{subsec:preliminary_study}. It shows that a fixed temperature can't adequately meet our needs.}
    \label{fig:temp_distribution}
\end{figure}

When performing downstream generation tasks, the attention is paid not only on the quality of the output but also on factors such as diversity \citep{chung2023increasing},  factual consistency \citep{tam2022evaluating}, etc. The factors influencing LLMs' performance in these aspects are highly complex. There are many reasons why we make great efforts to enhance the model's performance in these aspects, especially considering that optimization for these metrics can be crucial in certain scenarios. Taking diversity as an example, at times we may choose a powerful LLM as an oracle to generate more diversified content we require for a specific task while still ensuring high quality \citep{sultan2020importance}. Not to mention, the model's current generation may not always be satisfactory, in which cases multiple regenerations are needed, making it unacceptable if the model generates highly similar or even identical content every time.

To achieve control over the decoding process and the model generation, temperature sampling \citep{ackley1985learning}, one of the most commonly used sampling control methods, is always utilized during the decoding process, which influences the model performance by adjusting the probability distribution of the next token to be generated. However, fixed temperature settings are predominantly employed in present~\citep{ouyang2023llm, chiang2023can}, which still has a lot of shortcomings. To illustrate this point more concretely, we analyze the selection of temperature for optimal generation quality at the single instance level for 1000 instances on four different tasks (details are elaborated in the \S \ref{subsec:preliminary_study}). The results are shown in Figure \ref{fig:temp_distribution}. It is evident that any fixed temperature will not be the best option in a considerable number of cases no matter what kind of language tasks the model performs.

While temperature sampling strategy has also been widely used to strike a balance between generation quality and diversity~\citep{nasir2023llmatic}, we have observed significant limitations in fixed temperature settings. Given this premise, we should and can find a better, lighter and reasonable strategy for dynamically selecting the temperature. It is worth noting that the issue of how to appropriately select dynamic temperature during LLMs' decoding process has also caught more attention from other researchers~\citep{chang2023kl, zhu2023improving}, which also provides more confidence and inspiration for us. However, these existing relevant works propose some strategies intuitively, while still having some limitations in their methods and lacking comprehensive analysis of the relationship between their strategies and behaviours of LLMs. A comprehensive analysis is important for us to better understand the impact of temperature on LLMs' decoding process and generation.

In this work, we conduct sufficient investigation and analysis about the influences of temperature parameter on LLMs' generation. And based on these we propose a novel token-level dynamic entropy-based temperature sampling algorithm called \method, which can dynamically select the temperature at the current decoding step. Most importantly, we also comprehensively evaluate our algorithm's performance through some corresponding coordinate plots and some composite metrics based on generation quality and diversity introduced in \S \ref{subsec:experimental_setup}, which shows that our algorithm has achieved significant improvements compared with the baseline strategies while incurring nearly negligible computational costs. And our algorithm has approximately the same cost as the fixed temperature one, saving approximately half of the GPU memory compared to the other dynamic temperature sampling algorithm.

\section{Background}\label{sec:background}
In this section, we briefly introduce the background. We first overview the Large Language Models~(LLMs) and the basic paradigms while applying LLMs in many natural language generation tasks~($\S$\ref{ss:LLM4NLG}). 
Then we introduce the advanced sampling technologies in~$\S$\ref{kl_dynamic_temperature_sampling}.
\subsection{Large Language Models for Natural Language Generation}\label{ss:LLM4NLG}

With a much larger number of parameters pre-trained on a large corpora, large language models have shown the impressive language capabilities in a variety of language tasks and scenarios \citep{{pang2022long}, zou2023representation}. Given an input $X = (x_1,x_2,\cdots,x_{m-1}, x_m)$, the basic paradigm for LLMs to predict the output sequence $Y=(y_1,y_2,\cdots,y_{n-1},y_n)$ is:
\begin{equation}
    p(Y\vert X) = \prod^n_{t=1}p(y_t\vert y_{<t}, X)
\end{equation}

Based on this, sampling-based methods are proposed to randomly select the next token based on the probability distribution to enhance the randomness and diversity during generation \citep{zhao2023survey}:
\begin{equation}
    y_t \sim p(y\vert y_{<t}, X)
\end{equation}

Natural Language Generation (NLG) is a collection of a wide range of generative language tasks, which aims to generate text content in a specific task context. In recent years, many models and methods have been proposed for NLG tasks~\citep{li2017paraphrase, joshi2020spanbert}. Especially the LLMs have demonstrated remarkable capabilities across various NLG tasks~\citep{brown2020language, chowdhery2023palm, touvron2023llama}, ushering the research of NLG in a new phase. This lead to a new paradigm for NLG tasks called ``pre-train + prompt + predict''. 

We hope to enhance the ability of LLMs to get more diversity in many kinds of language tasks while maintaining good generation quality, which is very important and is attracting more attention. For summarization task, diverse summarization statements for a long piece of text can offer us more perspectives \citep{aralikatte2021focus}. As to question answering, it's a highly diverse task type itself, containing community question answering~\citep{li2022community}, conversational question answering \citep{zhu2018sdnet}, knowledge-based question answering \citep{chen2019bidirectional}, visual question answering~\citep{kazemi2017show} and so on. A better performance in terms of both generation quality and diversity seems to be important considering that the high diversity of answers in the real world \citep{nie2022unsupervised} and the single answer generated by LLMs may not always align with what users desire, which makes diversity become especially crucial. LLMs has pushed language models to new heights in machine translation~\citep{zhu2023multilingual} too. And the problems in improving the lexical diversity of generated translation have also be widely studied \citep{vanmassenhove2019lost, gu2020token}.

\subsection{Dynamic Temperature Sampling}\label{kl_dynamic_temperature_sampling}



\citet{ackley1985learning} is the first to introduce a temperature sampling strategy to adjust the probability distribution in a sampling-based decoding strategy. 
Given a probability distribution $p$ and a parameter $T$, it computes the sampling probability for sampling $k$-th choice with:
\begin{equation}
    p(t_k) = \frac{{\rm exp}(l_k / T)}{\sum_i {\rm exp}(l_i / T)}
\end{equation}
where $t_k$ means the $k$-th token in vocabulary, $l_i$ means the corresponding logits value of the $i$-th token, $T$ is the pre-specified temperature parameter. 

Temperature sampling is a common decoding strategy for LLMs’ generation process control, in which a higher temperature always leads to a more creative generation while a lower temperature leading to a more high-quality generation but with less variation in most cases. Due to the significant impact of temperature selection on the model's generation results, there have been work \citep{chang2023kl} attempting to achieve a better trade-off between diversity and attribution by dynamically selecting temperature. They use two parallel models to decode simultaneously, and select $T$ as temperature for the token to be generated in this step according to the KL-divergence of distribution. 
The expression for the selected $T$ is:
\begin{equation}\label{KLD_Equation}
    T = T_0 \cdot (\frac{1}{2})^{\frac{\operatorname{KL}(p||q)}{\sigma}}
\end{equation}
where $T_0$ is the baseline temperature and $\sigma$ is a hyperparameter to specify the half-life cycle of the decay. 

However, there are still some limitations in this method. Using two parallel models means much more GPU memory usage and some inherent limitations of distributed system, which imply instability and higher hardware requirements.

\section{Approach}\label{sec:entropy-based_dynamic_temperature_sampling}
In this section, we propose \textbf{E}ntropy-based \textbf{D}ynamic \textbf{T}emperature~(\method), a new temperature-selecting strategy. 
We firstly analyze the temperature distribution as our motivation for dynamic temperature sampling (\S \ref{subsec:preliminary_study}). 
We then introduce our novel paradigm to NLG that controls the LLMs' decoding based on model confidence (\S \ref{subsec:model_confidence_intro} and \S \ref{subsec:EDT_Sampling}), which can dynamically select the temperature at every decoding step for LLMs in a task-agnostic manner that is easy to deploy.

\subsection{Preliminary Study}\label{subsec:preliminary_study}
It's widely known that the fixed temperature algorithm is most frequently used in the systems involving temperature sampling. To elaborate its shortcomings and the necessity of dynamic temperature selecting, we analyze the optimal temperature under the same four different benchmarks as Section \ref{sec:experiments}, including XLSum~\citep{hasan2021xl}, MS MARCO v1.1~\citep{bajaj2016ms}, QuAC~\citep{choi-etal-2018-quac}, and WMT19. Following the basic experiment settings in \S \ref{subsec:experimental_setup}, we obtain the best temperature for every instance in these benchmarks and report statistical results in Figure \ref{fig:temp_distribution}. The results depicted in the figure indicate that there is a better choice for a considerable number of instances when we employ a fixed temperature no matter what kind of language tasks the model is performing, which shows a necessary and promising research direction, dynamic temperature sampling.

Given the contextual relevance and differences between segments, LLMs exhibit significant fluctuations in confidence levels at different decoding steps. What we observe in our experiments and the statistics provided by~\citet{zhu2023improving} demonstrate it in natural language tasks. The relationship between uncertainty and generation quality of the model generation~\citep{lin2023generating} shows the opportunity to control the model's decoding process through model confidence.

\subsection{Model Confidence for Predicting}\label{subsec:model_confidence_intro}
We choose entropy as a metric for the confidence of the model in every decoding step. The larger the entropy is, the less confident we consider the model to be when selecting the current token, and the smaller the entropy is, the more confident we consider the model to be. 

The concept of entropy was first proposed by \citet{shannon1948mathematical} to measure the uncertainty of a system about its actual structure. The entropy of a n-state system was defined by Shannon as:
\begin{equation}\label{entropy_equation}
    {\rm Entropy} = -\sum_{i=1}^n p_i \operatorname{log}(p_i)
\end{equation}
where $p_i$ is the probability of occurrence of the $i$-th event.

Intuitively we feel that the gain in information from an event is inversely related to its probability of occurrence \citep{pal1991entropy}. Such a gain in information can be measured as:
\begin{equation}
    \Delta I = \operatorname{log} (1/p_i) = -\operatorname{log} (p_i)
\end{equation}

Based on these, we can measure the uncertainty of the model by the entropy of the probability distribution while predicting the current token. Higher entropy indicates more confusion, while lower entropy indicating higher confidence.
When the model is confused, considering that it is difficult for it to guarantee a appropriate choice at the current position, using a higher temperature can help the model explore more possible answers without significantly affecting output's quality. Conversely, when the model is very confident in current step, given LLMs' strong language processing abilities, we use a lower temperature to make the model more committed to its current decision, which is also helpful in addressing the long-tail problem in sampling-based generation strategies. During our work, we are delighted to find that the related work of \citet{zhu2023improving} in code generation has also validated our hypothesis in their domain.

\begin{figure}[tbp]
    \centering
    \includegraphics[width=0.5\textwidth]{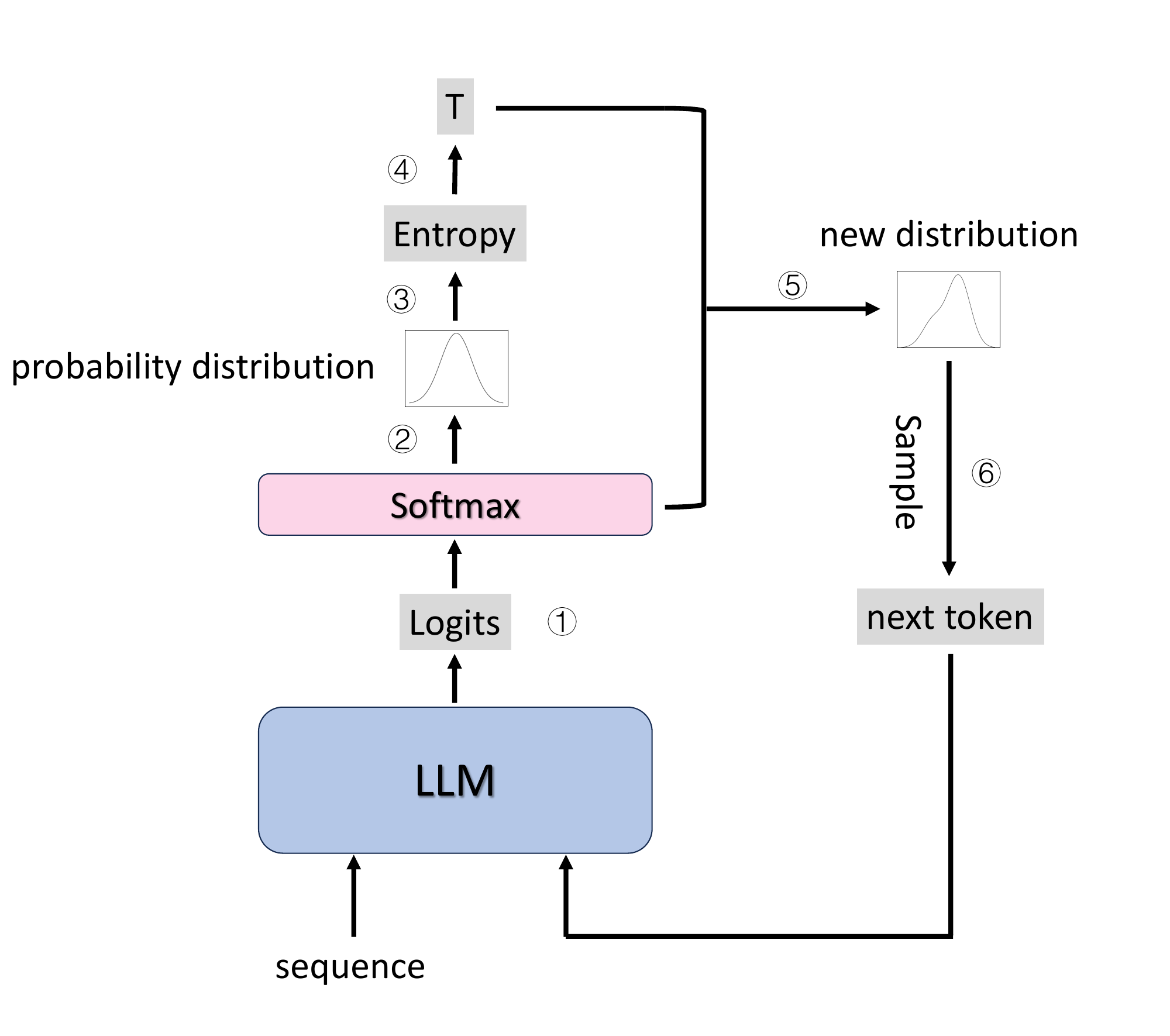}
    \caption{\textbf{Illustration of the decoding process with our \method}. At every decoding step, the system obtains the logits first (\ding{192}) and generate the probability distribution of the next token (\ding{193}). Then based on the entropy (\ding{194}) of the initial probability distribution, the model chooses the temperature (\ding{195}), obtains the new distribution (\ding{196}), and samples the next token (\ding{197}).}
    \label{fig:algorithm}
\end{figure}

\subsection{Entropy-based Temperature Selecting}\label{subsec:EDT_Sampling}
Following \citet{chang2023kl}, we aim to find a more lightweight and efficient method. Based on their work, in which they sample the temperature depending on the model decoder's prediction distribution, we propose a much lighter, simpler and more effective decoding strategy, Entropy-based Temperature Sampling. Unlike KL-divergence Guided Temperature Sampling algorithm (referred to as KLD algorithm below) that utilizes two parallel models for inference, we only use a single model and obtain the temperature through its prediction distribution at each step of the generation process. This means that we can save approximately half of the GPU memory usage, while also eliminating many potential bottlenecks (e.g. distributed system synchronization operations) in the two-parallel-decoding architecture.

We illustrate our algorithm process in Figure \ref{fig:algorithm}. While model is decoding, before generating the token finally, the logits and the token's prediction probability distribution are obtained by model first. According to our algorithm, we will calculate the entropy of the probability distribution first to measure the model's confidence in the current step according to the Eqn.~\ref{entropy_equation}. Then we can calculate the temperature for this step by
\begin{equation}\label{our_algorithm_equation}
    T = T_0\cdot \mathcal{N}^{\frac{\theta}{Entropy}},\ \ \ \ 0<\mathcal{N}<1 
\end{equation}
where $T_0$ and $\theta$ are both hyperparameters set before we employ the model for generation. $T_0$ indicates the upper bound of the temperature that we can sample throughout the process. $\theta$ affects the overall value size and scale of temperature variations. We set $\mathcal{N}=0.8$ in our all experiments below. In fact, the value of $\mathcal{N}$ also can be adjusted. The 0.8 is probably not the optimal choice in the majority of cases, but since we find it does not affect our demonstration of the effectiveness of our method, we simply choose 0.8 as a suitable value after some simple experimentation.

\paragraph{Parameter Tuning}
It is obvious that $T$ and $T_0$ are positively correlated, and for $T$ and $\theta$, we have:
\begin{equation}
    \frac{\delta T}{\delta \theta} = \frac{T_0\operatorname{ln}\mathcal{N}}{\rm Entropy}\cdot\mathcal{N}^{\frac{\theta}{\rm Entropy}}
\end{equation}
According to this equation, we can see that when $T_0 \neq 0$, i.e., in the case of non-greedy search, the derivative of $T$ with respect to $\theta$ is always negative, which means $T$ is a monotonic function with respect to $\theta$. 

This provides us with a straightforward parameter tuning approach, where we can start with a basic set of hyperparameters, such as $T_0=0.6$ and $\theta=0.1$, and then adjust $T_0$ or $\theta$ unidirectionally based on our requirements with the other hyperparameter fixed. Additionally, since the temperature values sampled at a given $T_0$ are always less than $T_0$, some infeasible $T_0$ values can be directly ruled out before parameter tuning according to practical requirements.

Finally, we can calculate the new prediction probability distribution using logits and obtain the temperature. After generating the current token, the system proceeds to the next iteration as usual, continuing to predict the next token.

\section{Experiments}\label{sec:experiments}
We evaluate our proposed dynamic temperature selecting strategy on several representative benchmarks, including text summarization, question answering, and machine translation. 
\subsection{Experimental Setup}\label{subsec:experimental_setup}
\paragraph{Datasets.}\label{tasks_and_datasets} 
We choose representative benchmarks for each task:
\begin{itemize}
    \item \textbf{Summarization}: We use the XLSum~\citep{hasan2021xl} benchmark to evaluate the text summarization task. 
    Specifically, we extract $10$k instances from the training subset of the XLSum English dataset as the training set. 
    And we randomly extract $1$k instances from the test subset for testing. 
    \item \textbf{Question Answering}: We conduct question answering~(QA) task on the QuAC~\citep{choi-etal-2018-quac} and MS MARCO v1.1~\citep{bajaj2016ms} datasets. 
    In detail, we extract $10$k instances from each training set for training and $1$k instances from the validation set for testing, respectively.
    \item \textbf{Translation}: We select the validation subset of WMT19 English-to-Chinese dataset\footnote{\url{https://statmt.org/wmt19/}} for evaluating the machine translation. Samely, we use $10$k instances for training and $1$k instances for testing.
\end{itemize}
More details about standardizing data and task prompting can be found in Appendix~\ref{sec:appendix_experiment_settings}.

\paragraph{Baselines.}
We include two mainstream temperature selecting strategies for comparison:
\begin{itemize}
    \item \textbf{Fixed}: Using the pre-defined temperatures while decoding. And we set the temperature $T$ from $0.1$ to $1.0$ which is commonly used.
    \item \textbf{Dynamic}: KLD~\citep{chang2023kl}, which uses the dynamic temperatures based on KL-divergence between the distribution of the conditional and unconditional decoding mode.
\end{itemize}

\begin{figure*}[tbp]
    \centering
    \begin{minipage}[b]{0.39\textwidth}
    \includegraphics[width=0.95\linewidth]{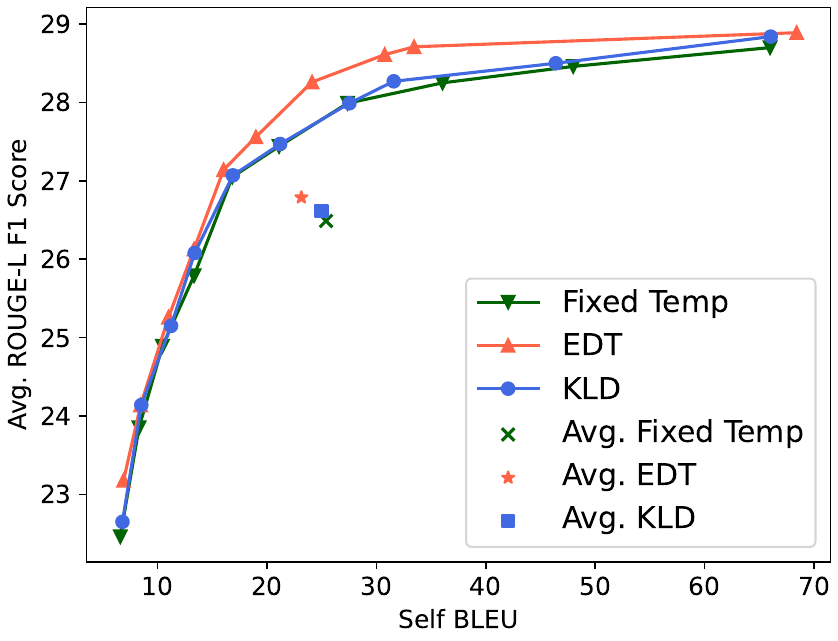}
    \subcaption{XLSum}
    \label{fig:summarization}
    \end{minipage}
    \hspace{15mm}
    \begin{minipage}[b]{0.40\textwidth}
    \includegraphics[width=0.95\linewidth]{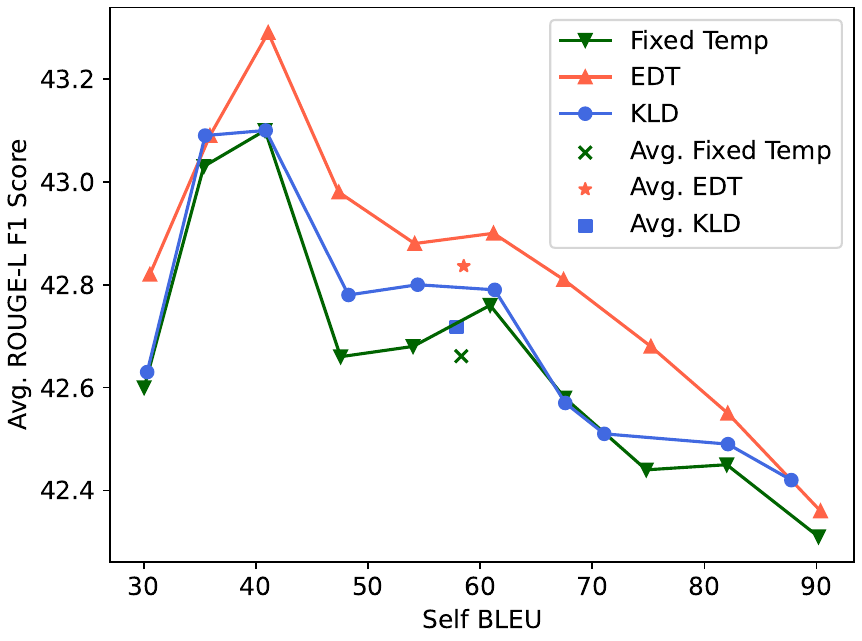}
    \subcaption{QuAC}
     \label{fig:qa_quac}
    \end{minipage}
    \begin{minipage}[b]{0.39\textwidth}
    \includegraphics[width=0.95\linewidth]{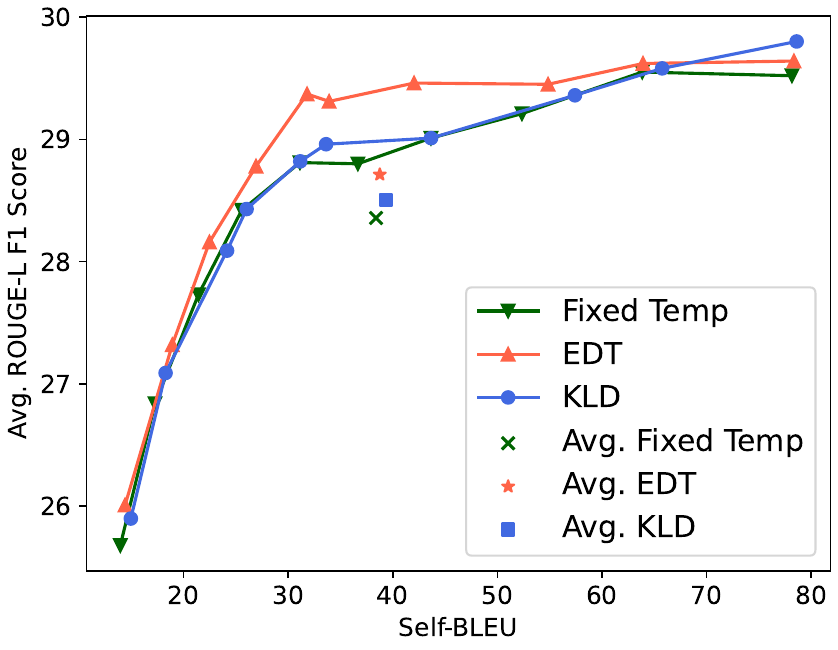}
    \subcaption{MS MARCO v1.1}\label{fig:qa_msmarco}
    \end{minipage}
    \hspace{15mm}
    \begin{minipage}[b]{0.39\textwidth}
    \includegraphics[width=0.95\linewidth]{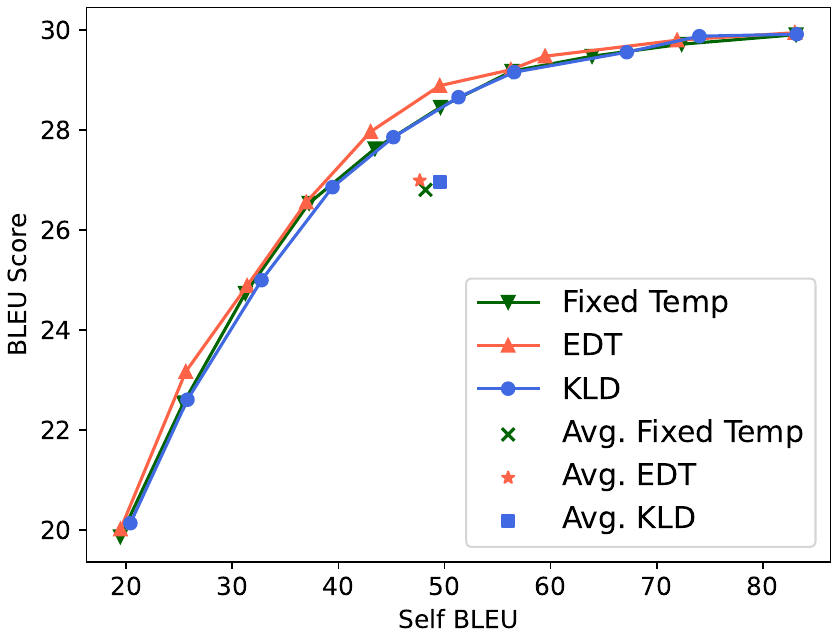}
    \subcaption{WMT19}\label{fig:translation}
    \end{minipage}
\caption{\textbf{Quality score~(higher is better) and diversity score~(lower is better) of sampling with different temperature strategies on different benchmarks}. We also plot the ``Avg. Fixed Temp'', ``Avg. EDT'' and ``Avg. KLD'' to show the average performances of ``Fixed Temp'', ``EDT'', and ``KLD'', correspondingly. The upper-left corner~(larger quality score but smaller diversity score) indicates a better performance.}
\label{fig:main_results}
\end{figure*}

\paragraph{Metrics.}
We evaluate model performances from quality and diversity with the following metrics:
\begin{itemize}
    \item \textbf{ROUGE-L and BLEU}: For the summarization and question-answering task, we evaluate the quality with the average F1 score of ROUGE-L~\citep{lin2004rouge} between the reference and sampled outputs, following \citet{aharoni2022mface} and \citet{nishida2019multi}. For the translation task, we use the average SacreBLEU~\citep{post2018call} score to evaluate the generation quality.
    \item \textbf{Self-BLEU}: We use the average Self-BLEU score between the sampled outputs to measure the generation diversity, following \citet{zhu2018texygen}.
\item \textbf{EDA}: As the tradeoff between generation quality and diversity always exists, we refer to \citet{li2021mixup} and compute the EDA~(\textbf{E}uclidean \textbf{D}istance from the ultimate \textbf{A}im) score to reflect the comprehensive performance as:
\begin{equation}\label{EDA_equation}
    {\rm EDA} = 100\% * \sqrt{(\frac{\mathcal{Q}-{q}}{\mathcal{Q}})^2 + \omega^2(\frac{d}{\mathcal{D}})^2},
\end{equation}
where $q$ is the quality score evaluated by BLEU or ROUGE, $d$ is the diversity score evaluated by Self-BLEU, $\mathcal{Q}$ is the highest quality score, $\mathcal{D}$ is the highest diversity score, and $\omega = \frac{\mathcal{Q}}{\mathcal{D}}$ is the weight to balance the change scale between the two metrics.
\end{itemize}
In addition, we make a modification on Eqn.~\ref{EDA_equation} to show the re-normalized trade-off score as:
\begin{equation}\label{EDA_range_equation}
    {\rm EDA_{range}} = 100\% * \sqrt{(\frac{\mathcal{Q}-{q}}{\mathcal{Q} - q^*})^2 + (\frac{d^*-d}{\mathcal{D}-d^*})^2},
\end{equation}
where $q^*$ and $d^*$ are the lowest quality score and diversity score in our experiments, respectively. We change the lower bound of the original range of EDA from the theoretical 0 to the practical lower bound, for better comparing the performance differences between different methods.

\paragraph{Implementations.}
We fix the Top-p ($p=0.95$) following \citet{chang2023kl} during our experiments for dynamic strategy, and then we fix the base in the temperature sampling formula at 0.8. 
We build our algorithm based on the implementation of Meta in LLaMA 2\footnote{\url{https://github.com/facebookresearch/llama}}. And we fine-tune the LLaMA2-13B model on different datasets respectively first before we use it for our following experiment.

For all tasks we will investigate below, we all use LoRA \citep{hu2021lora} to fine-tune the pre-trained language models with 2 epochs, batch\_size=4, gradient\_accumulation\_steps=4, lr\_scheduler\_type=cosine based on the settings of LLaMA-Factory\footnote{\url{https://github.com/hiyouga/LLaMA-Factory}} \citep{llama-factory}, a widely used and recognized open-source project.

Notice that we use a different $\sigma$ in KLD from \citet{chang2023kl} as we experimentally find the original setting always has high self-BLEU scores~($>90$) in our tasks, which does not meet our goal for high-diverse results.
In detail, we expand $\sigma$ to $\{1E^{0},3E^{0},1E^{1},3E^{1},1E^{2},3E^{2},1E^{3},3E^{3}\}$ in our experiments, for adequately covering a range of diversity score values.
Besides, we extensively investigate their strategy within the same $T_0$ as ours to ensure consistency and fairness while comparing it with our method. 

To evaluate the model performance in our experiments, we obtain the quality score and the self-BLEU score of every instance first. Then we calculate the average of all quality scores as the final quality score on current task, and calculate the average of all self-BLEU scores as the ultimate diversity score.

\begin{table*}[tbp]
    \centering
    \small
    \begin{tabular}{lcccccccc}
    \toprule
    \multirow{2}*{\textbf{Methods}} & \multicolumn{2}{c}{\textbf{Summarization}} & \multicolumn{2}{c}{\textbf{QA(QuAC)}} & \multicolumn{2}{c}{\textbf{QA(MS MARCO)}} & \multicolumn{2}{c}{\textbf{Translation}}\\
    \cmidrule(lr){2-3}\cmidrule(lr){4-5}\cmidrule(lr){6-7}\cmidrule(lr){8-9}
    & $\rm EDA$ & $\rm EDA_{range}$ & $\rm EDA$ & $\rm EDA_{range}$ & $\rm EDA$ & $\rm EDA_{range}$ & $\rm EDA$ & $\rm EDA_{range}$ \\
    \midrule
    Fixed & 12.19 & 32.55 & \textbf{16.01} & 26.37 & 12.44 & 35.82 & 19.73 & 43.85 \\ \midrule
    KLD \citep{chang2023kl} & 12.18 & 32.35 & 16.15 & 22.31 & 12.66 & 35.69 & 19.94 & 43.86 \\
    EDT (ours) & \textbf{11.61} & \textbf{28.80} & 16.24 & \textbf{18.34} & \textbf{12.15} & \textbf{29.49} & \textbf{19.60} & \textbf{41.89} \\
    \bottomrule
    \end{tabular}
    \caption{\label{table:EDA}
    Best $\rm EDA$ and $\rm EDA_{range}$ scores of sampling with different temperature selection strategies on summarization, question answering, and translation benchmarks. We highlight the \textbf{best} results.}
\end{table*}

\subsection{Main Results}
Figure~\ref{fig:main_results} presents the scatter plot of the quality score and diversity score.
Clearly, we can see that \method is almost located in the upper-left corner of all counterparts, indicating that sampling with entropy-based temperatures achieves a better trade-off between generation quality and diversity. 
We summarize our empirical findings as follows:
\begin{itemize}
    \item[1.] \textbf{Decoding with appropriate temperatures is necessary.} In most generation tasks, different temperature parameters can bring significant changes in generation quality of LLMs. 
    As seen, it can result in the ROUGE score on XLSum from 23 to 29~(Figure~\ref{fig:summarization}), and the BLEU score on WMT19 from 20 to 30~(Figure~\ref{fig:translation}).
    Compared with sampling with fixed temperatures, using dynamic temperatures~(KLD and our \method) always can help LLM achieve better quality scores in all tasks. 
    \item[2.] \textbf{Entropy-based dynamic temperature selection is simple yet effective.} It can be seen that the average performance of EDT is always located in the upper left corner of the strong baseline~(KLD), indicating that it can better balance the trade-off between sampling quality and sampling diversity. At the same time, compared with KLD which requires two parallel decoding processes, our proposed EDT only requires a small amount of vector multiplications~(for computing entropy based on the decoding distribution), which hardly increases the inference cost, and is a simpler and more effective strategy.
\end{itemize}
Table~\ref{table:EDA} further shows the best $\rm EDA$ and $\rm EDA_{range}$ scores of the different temperature strategies.
It can be seen that $\rm EDA$ and $\rm EDA_{range}$ scores of EDT are better than those of its baselines, which shows that using entropy to guide the setting of temperature parameters helps the model better balance sampling quality and sampling diversity. This is also consistent with what we learned from Figure~\ref{fig:main_results} before.

\begin{figure}[tbp]
    \centering
    \includegraphics[width=0.4\textwidth]{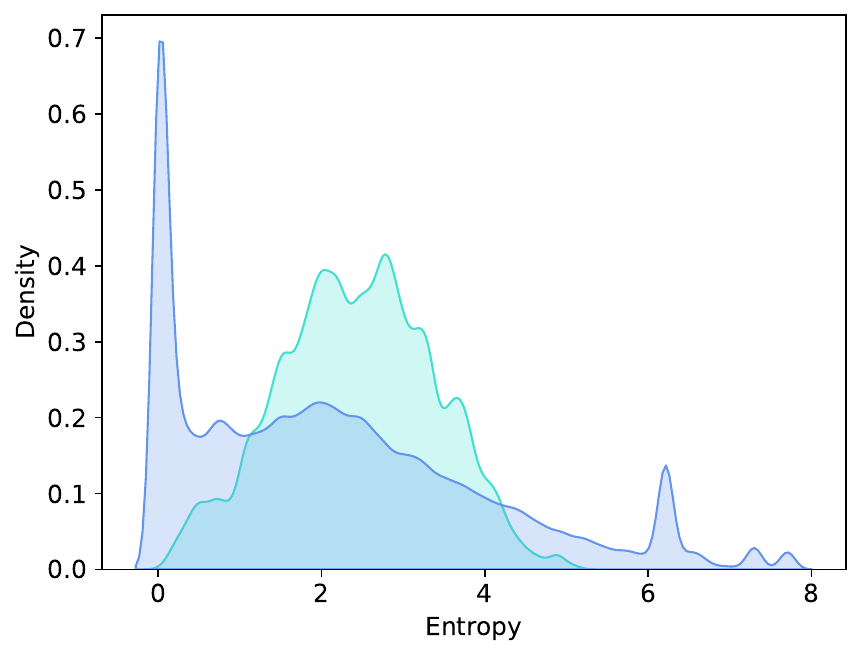}
    \caption{\textbf{Density distribution of entropy on XLSum.} For each example, we calculate the entropy of the distribution to decode the first token~(green) and all entropy of the distribution to decode all tokens~(blue), with teacher-forcing decoding mode.}
    \label{fig:entropy_density}
\end{figure}

\subsection{Analysis}

\paragraph{Token- vs. Instance-level Dynamic Temperatures.}
We further transfer our entropy-based dynamic temperatures to the instance level, which sets an instance-level sampling temperature based on the entropy of distribution to decode the first token. 
We evaluate its effectiveness and compare it to our existing token-level temperature selection strategy on summarization tasks.
We can see from Table~\ref{tab:EDA_instance} that using dynamic temperatures at the token level achieves better $\rm EDA$ and $\rm EDA_{range}$ scores, indicating that token-level \method allows for achieving better trade-offs.

To better understand the reason behind this, we calculate the density of different kinds of entropy in Figure~\ref{fig:entropy_density}. 
As seen, the entropy of the first token is usually higher, which is unable to generalize the model's decoding behavior across all tokens. 
In such a case, the instance-level \method degenerates to the fixed temperatures (but still dynamic).
Nevertheless, we can see the performance of the instance-level \method is still better than that of fixed temperatures in terms of $\rm EDA_{range}$ score, demonstrating the necessity of changing the existing fixed temperature strategy for LLM decoding to dynamic temperatures.

\begin{table}[tbp]
    \centering
    \small
    \begin{tabular}{lcc}
        \toprule
        \textbf{Methods} & $\rm EDA^\downarrow$ & $\rm EDA_{range}^\downarrow$ \\
        \midrule
        Fixed& 12.19 & 32.55 \\\midrule
        Token-level EDT & \textbf{11.61} & \textbf{28.80} \\
        Instance-level EDT & 12.30 & 31.63 \\
        \bottomrule
    \end{tabular}
    \caption{Performances of sampling with token- and instance-level EDT on XLSum. To compare them directly, we report the best $\rm EDA$ and $\rm EDA_{range}$ scores of each temperature strategy.}
    \label{tab:EDA_instance}
\end{table}

\paragraph{Entropy- vs. Uncertainty-based Dynamic Temperatures.} 
We also analyze the effectiveness of using entropy as a measure of model confidence. 
To this end, we introduce an intuitive metric for evaluating the confidence of model prediction. 
In detail, we replace the $\rm Entropy$ in Eqn.~\ref{our_algorithm_equation} with
\begin{equation}
    {\rm Uncertainty} = \sqrt{1 - p_1}
\end{equation}
where $p_1$ is the top-1 probability in the distribution. Such \textbf{U}ncertainty-based \textbf{D}ynamic \textbf{T}emperatures~(denoted as UDT) can be obtained by
\begin{equation}
    T = T_0\cdot \mathcal{N}^{\frac{\theta}{\rm Uncertainty}}.
\end{equation}
We then conduct experiments on the question-answering task (MS MARCO benchmark) to evaluate its effectiveness. 
The results are listed in Table~\ref{tab:EDA_ablation_entropy}.

As shown, both UDT and \method show better performance than the fixed temperature strategy, which is consistent with our above observations. 
At the same time, we can easily notice that EDT has better performance than UDT, indicating that the entropy-based measurement strategy we chose is more suitable as a basis for regulating sampling temperature.


\begin{table}[tbp]
    \centering
    \small
    \begin{tabular}{lcc}
        \toprule
        \textbf{Methods} & $\rm EDA^\downarrow$ & $\rm EDA_{range}^\downarrow$ \\
        \midrule
        Fixed & 12.44 & 35.82 \\\midrule
        EDT & \textbf{12.15} & \textbf{29.49} \\
        UDT & 12.41 & 31.45 \\
        \bottomrule
    \end{tabular}
    \caption{Performances of sampling with entropy- and uncertainty-based dynamic temperature on MS MARCO.}
    \label{tab:EDA_ablation_entropy}
\end{table}

\paragraph{Effects of $T_0$ and $\theta$.} 
We conduct an ablation study in a question-answering task (MS MARCO benchmark) to analyze the impact of $T_0$ and $\theta$ in Eqn.~\ref{our_algorithm_equation}.
It is worth noting that $T_0$ determines the temperature range and $\theta$ plays an important role in the sensitivity of temperature to entropy. 
Experimental results in Table \ref{tab:ablation_T0_effects} and Table \ref{tab:ablation_theta_effects} verify this. 
First, as $T_0$ changes, the overall performance of the model will fluctuate accordingly, which reflects the impact of the overall temperature range on model performance.
When $T_0$ is fixed, controlling temperature changes through $\theta$ can help the model obtain better performance, but if $\theta$ is adjusted inappropriately, it will produce worse results than the original algorithm. 

Overall, we can see that setting appropriate hyperparameters $T_0$ and $\theta$ plays an important role in the effectiveness of our algorithm.


\begin{table}[tbp]
    \centering
    \resizebox{0.46\textwidth}{!}{
    \begin{tabular}{lcccccc}
        \toprule
        \textbf{$\theta$} & 0.1 & 0.3 & 0.5 & 1.0 & 3.0 & 5.0 \\
        \midrule
        $\rm EDA$ & 14.61 & 13.91 & 13.62 & \textbf{12.35} & 14.86 & 16.57 \\
        $\rm EDA_{range}$ & 92.72 & 84.98 & 80.19 & 62.29 & 54.90 & \textbf{42.10} \\
        \bottomrule
    \end{tabular}}
    \caption{Performances of sampling with fixed $T_0=1.0$ and different $\theta$ on MS MARCO.}
    \label{tab:ablation_T0_effects}
\end{table}

\paragraph{Case Study}
We conduct case study on XLSum dataset, whose results are shown in Table \ref{tab:case_study} in Appendix \ref{appendix:case_study}. We choose the best hyperparameter settings on $\rm EDA_{range}$ of every method in Figure \ref{fig:main_results} and follow the same experiment settings as Section \ref{sec:experiments}, generating five times for every instance. The outputs of EDT are obviously more succinct, while accurately conveying the original text's meaning. According to our results, the outputs of EDT achieve significantly better generation quality scores with similar self-BLEU. There is fewer redundant information in EDT’s outputs, which brings it a much higher ROUGE-L F1 score. In contrast, another two algorithms actually achieve higher diversity by incorporating more redundant background information in the answers.

\begin{table}[tbp]
    \centering
    \resizebox{0.46\textwidth}{!}{
    \begin{tabular}{lcccccc}
        \toprule
        \textbf{$T_0$} & 0.1 & 0.3 & 0.5 & 0.7 & 0.9 & 1.0 \\
        \midrule
        $\rm EDA$ & 37.78 & 25.37 & 17.94 & 13.25 & \textbf{12.72} & 14.61 \\
        $\rm EDA_{range}$ & 99.67 & 60.73 & 38.90 & \textbf{34.55} & 67.53 & 92.72 \\
        \bottomrule
    \end{tabular}}
    \caption{Performances of sampling with fixed $\theta=0.1$ and different $T_0$ on MS MARCO.}
    \label{tab:ablation_theta_effects}
\end{table}

\section{Conclusion}
In this paper, we present a novel paradigm in language generation tasks that dynamically adjusts LLM decoding behavior based on its confidence for predicting. Specifically, we propose an entropy-based dynamic temperature selection strategy, which chooses the temperature parameter for sampling. Experiments on several representative generation tasks validate it is simple enough to be seamlessly applied to a variety of language generation tasks and outperforms existing temperature sampling strategies. 
Our algorithm is simple enough to be seamlessly applied to a variety of generative language tasks, following the ``one-for-all'' spirit of LLM research. 
We hope our proposed LLM decoding strategy can inspire the followers to explore such a promising research direction.


\section*{Limitations}
Our goal is to draw attention to the study of dynamic temperature by proposing a simple and effective dynamic temperature sampling algorithm. Despite our method exhibits the anticipated effects across various NLG tasks, and demonstrates a significant improvement in both efficiency and effectiveness compared to existing methods, there is still some limitations waiting for research. Although our algorithm is task-agnostic, it is still limited by the specific tasks or data, which means the same set of hyperparameter settings can't be universally applied to all language tasks or datasets.

Furthermore, our method relies on certain manual configurations, implying that developing a neural network which is able to automatically selecting hyperparameters will be more efficient. In addition to that, a learnable network can even select hyperparameters for every instance, which can achieve a more effective control. It shows the learnable parameter selecting strategy is an important research direction.
\bibliography{custom}
\newpage
\appendix

\section{Details of Experiment Settings}\label{sec:appendix_experiment_settings}
We perform specific data processing operations for different tasks:
\begin{itemize}
    \item \textbf{Summarization: }We add "\textbackslash n" at the end of the model input during both training and inference to help the model learning this pattern and generating expected text summary.
    \item \textbf{Question Answering: }Considering that the input consists of context, the answer waiting to be answered, several history questions and their corresponding answers, we placed the context at the beginning of the input, followed by multiple historical question-answer pairs, and at the end of the input is our question. It's worth noting that we also added "\textbackslash n\textbackslash nQ: " before each question and "\textbackslash nA: " before each answer, and appended "\textbackslash nA: " at the end of the input to help the model generate answer more effectively.
    \item \textbf{Translation: }We add "\textbackslash nTranslate English to Chinese:\textbackslash n" at the end of each input not only during fine-tuning but also during inference.
\end{itemize}

\section{Case Study}\label{appendix:case_study}
Results are shown in Table \ref{tab:case_study}.
\begin{table*}[tbp]
    \centering
    \small
    \renewcommand\arraystretch{1.2}
    \begin{tabular}{p{15cm}}
        \hline
        \multicolumn{1}{c}{\textbf{Text}} \\
        \hline
        In 2016, when ex-President Dilma Rousseff was accused of manipulating the government budget and her impeachment trial was launched, then Vice-President Temer stepped in as acting president. As BBC South America business correspondent Daniel Gallas wrote at the time, Mr Temer, had "always been a kingmaker, but never king". \textbf{$\cdots\cdots$} \\
        \hline
        \multicolumn{1}{c}{\textbf{Reference Output}} \\
        \hline
        + For many Brazilians, Michel Temer's ascension to president was surprising. But as the first Brazilian president to have charges brought against him while in office, his arrest is less unexpected. \\
        \hline
        \multicolumn{1}{c}{\textbf{Summary} (Fixed Temperature T = 0.5)} \\
        \hline
        + Michel Temer's presidency was short-lived but it was also controversial. \\
        + Michel Temer, Brazil's former president, has been charged with corruption by federal prosecutors. He is accused of leading a bribery scheme when he was vice-president and then president. \\
        + Michel Temer, Brazil's former president, was sentenced to nine years and six months in prison for corruption on Thursday. BBC Brasil's Daniel Gallas looks at how the former vice-president rose to power and what the ruling means for him. \\
        + Michel Temer, who has resigned as Brazil's president, was a key figure in the country's political elite for decades. He was the vice-president under Dilma Rousseff and became acting president when she was impeached. He was the first sitting president in Brazil to be charged with a crime. \\
        + Michel Temer, who has died aged 77, was the first Brazilian president to be charged with corruption. He was also the first to be accused of trying to obstruct justice. He was also accused of running a bribery scheme that funnelled hundreds of millions of dollars from state-run institutions into the pockets of lawmakers. \\
        \textcolor{red}{ROUGE-L\ F1: 15.00 \ \ \ self\_BLEU: 5.43} \\
        \hline
        \multicolumn{1}{c}{\textbf{Summary} (KLD $(T_0, \delta)$ = (0.5, 3e2))} \\
        \hline
        + Michel Temer's presidency was short-lived but it was also controversial. \\
        + Michel Temer, Brazil's former president, has been charged with corruption by federal prosecutors. He is accused of leading a bribery scheme when he was vice-president and then president. \\
        + Michel Temer, Brazil's former president, was sentenced to nine years and six months in prison for corruption on Thursday. BBC Brasil's Daniel Gallas looks at how the former vice-president rose to power and what the ruling means for him. \\
        + Michel Temer, who has resigned as Brazil's president, was a key figure in the country's political elite for decades. He was the vice-president under Dilma Rousseff and became acting president when she was impeached. He was the first sitting president in Brazil to be charged with a crime. \\
        + Michel Temer, who has died aged 77, was the first Brazilian president to be charged with corruption. He was also the first to be accused of trying to obstruct justice. He was also accused of running a bribery scheme that funnelled hundreds of millions of dollars from state-run institutions into the pockets of lawmakers. \\
        \textcolor{red}{ROUGE-L\ F1: 15.00 \ \ \ self\_BLEU: 5.43} \\
        \hline
        \multicolumn{1}{c}{\textbf{Summary} (EDT $(T_0, \theta)$ = (0.7, 2e1))} \\
        \hline
        + Michel Temer, Brazil's former president, has been charged with corruption and money laundering. \\
        + Michel Temer, who has died aged 77, was an unpopular figure who took over as Brazil's president in 2016, following the impeachment of Dilma Rousseff. \\
        + Michel Temer, who has resigned as Brazil's president, was a career politician who was unpopular with the public and plagued by corruption allegations. \\
        + Michel Temer, the former Brazilian president, has been sentenced to nine years in prison for corruption. He is the first Brazilian president to be found guilty of corruption. \\
        + Michel Temer has stepped down as Brazil's president after the Supreme Court rejected his appeal against a 10-year ban on holding public office. He was accused of corruption and had been convicted by the lower courts. He was the first Brazilian president to be charged with a crime while in office. \\
        \textcolor{red}{ROUGE-L\ F1: 24.69 \ \ \ self\_BLEU: 7.44} \\
        \hline
    \end{tabular}
    \caption{We illustrate the advantages of our method in this case. We choose the best hyperparameter settings of every method in Figure \ref{fig:main_results}. While the outputs of these three methods have close self-BLEU, there is fewer redundant information in EDT's outputs, which brings it a much higher ROUGE-L F1 score.}
    \label{tab:case_study}
\end{table*}

\end{document}